\documentclass[runningheads,a4paper]{llncs}
\usepackage{hyperref}
\usepackage{amssymb}
\usepackage{graphicx,epstopdf,epsfig,subcaption,amsmath}
\usepackage{tikz}
\usepackage{setspace}
\usepackage{color}
\usepackage{hhline}
\usepackage{url}
\usepackage{multirow}
\usepackage{array,booktabs}
\urldef{\mailsa}\path|{alfred.hofmann, ursula.barth, ingrid.haas, frank.holzwarth,|
\urldef{\mailsb}\path|anna.kramer, leonie.kunz, christine.reiss, nicole.sator,|
\urldef{\mailsc}\path|erika.siebert-cole, peter.strasser, lncs}@springer.com|

\usepackage{floatrow}
\newfloatcommand{capbtabbox}{table}[][\FBwidth]

\newsavebox\CBox
\def\textBF#1{\sbox\CBox{#1}\resizebox{\wd\CBox}{\ht\CBox}{\textbf{#1}}}
\begin{document}
\mainmatter  

\makeatletter
\def\@normalsize{\@setsize\normalsize{10pt}\xpt\@xpt
\abovedisplayskip 10pt plus2pt minus5pt\belowdisplayskip
\abovedisplayskip \abovedisplayshortskip \z@
plus3pt\belowdisplayshortskip 6pt plus3pt
minus3pt\let\@listi\@listI}
\def\subsize{\@setsize\subsize{12pt}\xipt\@xipt}
\def\section{\@startsection {section}{1}{\z@}{1.0ex plus
1ex minus .2ex}{.2ex plus .2ex}{\large\bf}}
\def\subsection{\@startsection {subsection}{2}{\z@}{.2ex
plus 1ex} {.2ex plus .2ex}{\subsize\bf}} \makeatother

\title{Improving Deep Pancreas Segmentation in CT and MRI Images via Recurrent Neural Contextual Learning and Direct Loss Function}
\author{Jinzheng Cai$^1$, Le Lu$^3$, Yuanpu Xie$^1$, Fuyong Xing$^2$, Lin Yang$^{1,2}$}
\institute{
$^1$ Department of Biomedical Engineering, \\
$^2$ Department of Electrical and Computer Engineering, \\ 
University of Florida, Gainesville, FL 32611, USA \\
$^3$ Department of Radiology and Imaging Sciences,\\
National Institutes of Health Clinical Center, Bethesda, MD 20892, USA}
\toctitle{Lecture Notes in Computer Science}
\tocauthor{Authors' Instructions}
\titlerunning{Pancreas Segmentation in MRI using Graph-Based Decision Fusion on  CNN}
\authorrunning{J. Cai et al.}

\maketitle

\begin{abstract}
Deep neural networks have demonstrated very promising performance on accurate segmentation of challenging organs ({\it e.g.,} pancreas) in abdominal CT and MRI scans. The current deep learning approaches conduct pancreas segmentation by processing sequences of 2D image slices independently through deep, dense per-pixel masking for each image, without explicitly enforcing spatial consistency constraint on segmentation of successive slices. We propose a new convolutional/recurrent neural  network architecture to address the contextual learning and segmentation consistency problem. A deep convolutional sub-network is first designed and pre-trained from scratch. The output layer of this network module is then connected to recurrent layers and can be fine-tuned for contextual learning, in an end-to-end manner. Our recurrent sub-network is a type of Long short-term memory (LSTM) network that performs segmentation on an image by integrating its neighboring slice segmentation predictions, in the form of a dependent sequence processing. Additionally, a novel segmentation-direct loss function (named Jaccard Loss) is proposed and deep networks are trained to optimize Jaccard Index (JI) directly. Extensive experiments are conducted to validate our proposed deep models, on quantitative pancreas segmentation using both CT and MRI scans. Our method outperforms the state-of-the-art work on CT \cite{DBLP:journals/corr/RothLFSS16} and MRI pancreas segmentation \cite{Cai2016}, respectively.      
\end{abstract}

\section{Introduction}
Detecting unusual volume changes and monitoring abnormal growths in pancreas using medical images is a critical yet challenging diagnosis task. This would require to dissect pancreas from its surrounding tissues in radiology images ({\it e.g.,} CT and MRI scans). Manual pancreas segmentation is laborious, tedious, and sometimes prone to inter-observer variability. One major group of related work on automatic pancreas segmentation in CT images are based on multi-atlas registration and label fusion (MALF) \cite{Oda2016,Tong201592,wolz2013automated} under leave-one-patient-out evaluation protocol. Due to the high deformable shape and vague boundaries of pancreas in CT, their reported segmentation accuracy results (measured in Dice Similarity Coefficient or DSC) range from 69.6$\pm$16.7\% \cite{wolz2013automated} to 75.1$\pm$15.4\% \cite{Oda2016}. On the other hand, deep convolutional neural networks (CNN) based pancreas segmentation work \cite{Cai2016,DBLP:journals/corr/FaragLRLTS15,roth2015deeporgan,DBLP:journals/corr/RothLFSS16,DBLP:journals/corr/ZhouXSFY16,Roth2017} have revealed promising results and steady performance improvements, e.g., from 71.8$\pm$10.7\% \cite{roth2015deeporgan}, 78.0$\pm$8.2\% \cite{DBLP:journals/corr/RothLFSS16}, to 81.3$\pm$6.3\% \cite{Roth2017} evaluated using the same NIH 82-patient CT dataset \url{https://doi.org/10.7937/K9/TCIA.2016.TNB1KQBU}.

In comparison, deep CNN approaches appear to demonstrate the noticeably higher segmentation accuracy and numerically more stable results (significantly lower in standard deviation, or std) than their MALF counterparts. \cite{DBLP:journals/corr/RothLFSS16,Roth2017} are built upon the fully convolutional network (FCN) architecture\cite{long2014fully} and its variant\cite{xie2015holistically}. However, \cite{DBLP:journals/corr/RothLFSS16,Roth2017} are not completely end-to-end trained due to their segmentation post processing steps. Consequently, the trained models may be suboptimal. For pancreas segmentation on a 79-patient MRI dataset, \cite{Cai2016} achieves 76.1$\pm$8.7\% in DSC. 

In this paper, we propose a new deep neural network architecture with recurrent neural contextual learning for improved pancreas segmentation. All previous work \cite{Cai2016,DBLP:journals/corr/RothLFSS16,DBLP:journals/corr/ZhouXSFY16} perform deep 2D CNN segmentation on either CT or MRI image or slice independently\footnote{Organ segmentation in 3D CT and MRI scans can also be performed by directly taking cropped 3D sub-volumes as input \cite{DBLP:journals/corr/MilletariNA16,DBLP:journals/corr/MerkowKMT16,DBLP:journals/corr/KamnitsasLNSKMR16}. Even at the expense of being computationally expensive and prone-to-overfitting, the result of very high segmentation accuracy has not been reported for complexly shaped organs \cite{DBLP:journals/corr/MerkowKMT16}. \cite{DBLP:journals/corr/ChenYZAC16,DBLP:journals/corr/StollengaBLS15} use hybrid CNN-RNN architectures to process/segment sliced CT or MRI images in sequence.}. There is no spatial smoothness consistency constraints enforced among successive slices. We first follow this protocol by training 2D slice based CNN models for pancreas segmentation. Once this step of CNN training converges, inspired by sequence modeling for precipitation nowcasting in \cite{DBLP:journals/corr/ShiCWYWW15}, a convolutional long short-term memory (CLSTM) network is further added to the output layer of the deep CNN to explicitly capture and constrain the contextual segmentation smoothness across neighboring image slices. Then the whole integrated CLSTM network can be end-to-end fine-tuned via stochastic gradient descent (SGD) until converges. The CLSTM module will modify the segmentation results produced formerly by CNN alone, by taking the initial CNN segmentation results of successive axial slices (in either superior or interior direction) into account. Therefore the final segmented pancreas shape is constrained to be consistent among adjacent slices, as a good trade-off between 2D and 3D segmentation deep models.

Next, we present a novel segmentation-direct loss function to train our CNN models by minimizing the jaccard index between any annotated pancreas mask and its corresponding output segmentation mask. The standard practice in FCN image segmentation deep models \cite{long2014fully,xie2015holistically,Cai2016,DBLP:journals/corr/RothLFSS16} use a loss function to sum up the cross-entropy loss at each voxel or pixel. Segmentation-direct loss function can avoid the data balancing issue during CNN training between the positive pancreas and negative background regions. Pancreas normally only occupies a very small fraction on each slice. Furthermore, there is no need to calibrate the optimal probability threshold to achieve the best possible binary pancreas segmentation results from the FCN's probabilistic outputs \cite{long2014fully,xie2015holistically,Cai2016,DBLP:journals/corr/RothLFSS16}. Similar segmentation metric based loss functions based on DSC are concurrently proposed and investigated in \cite{DBLP:journals/corr/MilletariNA16,DBLP:journals/corr/ZhouXSFY16}. 

We extensively and quantitatively evaluate our proposed deep convolutional LSTM neural network pancreas segmentation model and its ablated variants using both a CT (82 patients) and one MRI (79 patients) dataset, under 4-fold cross-validation (CV). Our complete model outperforms $4\%$ of DSC comparing to previous state-of-the-arts\cite{Cai2016,DBLP:journals/corr/RothLFSS16}. Although our contextual learning model is only tested on pancreas segmentation, the approach is directly generalizable to other three dimensional organ segmentation tasks. 

\section{Method} \label{section:method}
\begin{figure*}[t!]
	\centering
	\includegraphics[width=12cm, height=3cm]{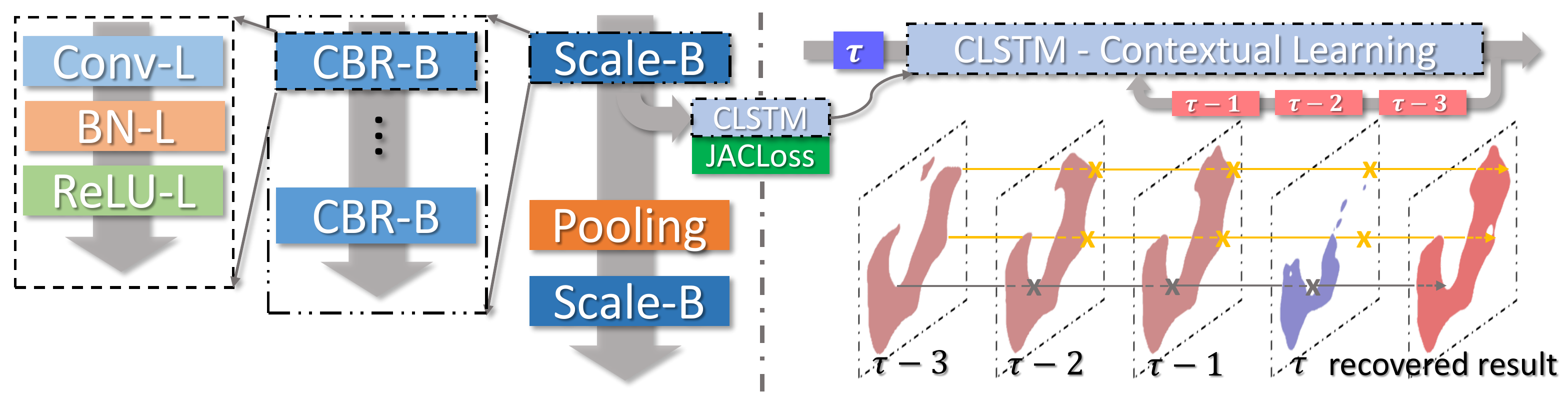}
	\caption{\small {\bf Network architecture:} Left is the CBR block (CBR-B) that contains convolutional layer (Conv-L), batch normalization layer (BN-L), and ReLU layer (ReLU-L). While, each scale block (Scale-B) has several CBR blocks and followed with a pooling layer. Right is the CLSTM for contextual learning. Segmented outcome at slice $\tau$ would be regularized by the results of slice $\tau-3$, $\tau-2$, and $\tau-1$. For example, contextual learning is activated in regions with $\times$ markers, where sudden losses of pancreas areas occurs in slice $\tau$ comparing to consecutive slices.}
	\label{fig:architecture}
\end{figure*}

{\bf Simplifying Deep CNN Architecture:} We propose to train deep CNN network from scratch and empirically observe that, for the specific application of pancreas segmentation in CT/MRI images, ImageNet pre-trained CNN models do not noticeably improve the performance. More importantly, we design our CNN network architecture specifically for pancreas segmentation where a much smaller CNN model than the conventional models \cite{xie2015holistically,long2014fully}is found to be most effective. This model reduces the chance of over-fitting (against small-sized medical image datasets) and can speed up both training and inference. Our specialized deep network architecture is trained from scratch using pancreas segmentation datasets, without being first pre-trained using ImageNet \cite{xie2015holistically,long2014fully} and then fine-tuned. It also outperforms the ImageNet fine-tuned conventional CNN models \cite{xie2015holistically,long2014fully} from our empirical evaluation. 

First, {\it Convolutional} layer is followed by {\it ReLU} and {\it Batch normalization} layers to form the basic unit of our customized network, namely the CBR block. Second, following the deep supervision principle proposed in \cite{xie2015holistically}, we stack several CBR blocks together with an auxiliary loss branch per block and denote this combination as Scale block. Fig.\ref{fig:architecture} shows exemplar CBR block (CBR-B) and Scale block (Scale-B). Third, we use CBR block and Scale block as the building blocks to construct our tailored deep network, with each Scale block is followed with a pooling layer. Hyper parameters of the numbers of feature maps in convolutional layers, the number of CBR blocks in a Scale block, as well as the number of Scale blocks to fit into our network can be determined via a model selection process on a subset of training dataset (i.e., split for model validation). 

\subsection{Contextual Regularization}
\begin{figure*}[t!]
	\centering
	\includegraphics[width=12cm, height=2.3cm]{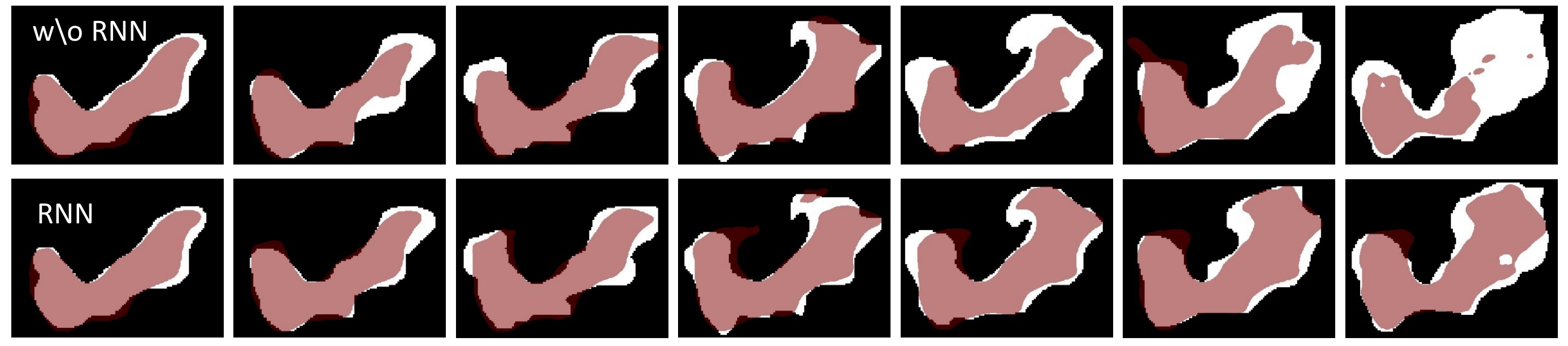}
	\caption{\small {\bf NIH Case51:} segmentation results with and without contextual learning are displayed in row 1 and row 2, respectively. Golden standards are displayed in white, and automatic outputs are rendered in red.}
	\label{fig:contextual-learning}
\end{figure*}

From above, we have designed a compact CNN architecture which can process pancreas segmentation on individual 2D image slices. However as shown in the first row of Fig.\ref{fig:contextual-learning}, the transition among the resulted CNN pancreas segmentation regions in consecutive slices may not be smooth, often implying that segmentation failure occurs. Adjacent CT/MRI slices are expected to be correlated to each other thus segmentation results from successive slices need to be constrained for shape consistence. 

To achieve this, we concatenate long short-term memory (LSTM) network to the 2D CNN model for contextual learning, as a compelling architecture for sequential data processing. That is, we slice any 3D CT (or MRI) volume into a 2D image sequence and process to learn the segmentation contextual constraints among neighboring image slices with LSTM. Standard LSTM network requires the vectorized input which would sacrifice the spatial information encoded in the output of CNN. We therefore utilize the convolutional LSTM (CLSTM) model \cite{DBLP:journals/corr/ShiCWYWW15} to preserve the 2D image segmentation layout by CNN. The second row of Fig.\ref{fig:contextual-learning} illustrates the improvement by enforcing CLSTM based segmentation contextual learning.

\subsection{Jaccard Loss} 
We propose a new jaccard loss (JACLoss) for training neural network image segmentation model. To optimize JI (a main segmentation metric) directly in network training makes the learning and inference procedures consistent and generate threshold-free segmentation. JACLoss is defined as follows: 
\begin{equation}
L_{jac} = 1 - \frac{|Y_+ \bigcap \hat{Y}_+|}{|Y_+ \bigcup \hat{Y}_+|} = 1 - \frac{\sum_{j \in Y} y_j \land \hat{y}_j}{\sum_{j \in Y} y_j \lor \hat{y}_j} = 1 - \frac{\sum_{f \in Y_+} (1 \land \hat{y}_f)}{|Y_+| + \sum_{b \in Y_-} (0 \lor \hat{y}_b)}
\end{equation}
where $Y$ and $\hat{Y}$ represent the ground truth and network predictions. Respectively, we have $Y_+$ and $Y_-$ defined as the foreground pixel set and the background pixel set, and $|Y_+|$ is the cardinality of $Y_+$. Similar definitions are also applied to $\hat{Y}$. $y_j$ and $\hat{y}_j \in \{0,1\}$ are indexed pixel values in $Y$ and $\hat{Y}$. In practice, $\hat{y}_j$ is relaxed to the probability number in range $[0,1]$ so that JACLoss can be approximated by 
\begin{equation}
\tilde{L}_{jac} = 1 - \frac{\sum_{f \in Y_+} \min(1,\hat{y}_f)}{|Y_+| + \sum_{b \in Y_-} \max(0,\hat{y}_b)}= 1 - \frac{\sum_{f \in Y_+}\hat{y}_f}{|Y_+| + \sum_{b \in Y_-}\hat{y}_b}
\end{equation}
Obviously, $L_{jac}$ and $\tilde{L}_{jac}$ are sharing the same optimal solution of $\hat{Y}$, with slight abuse of notation, we use $L_{jac}$ to denote both. The model is updated by:
\begin{equation}
\frac{\partial L_{jac}}{\partial \hat{y}_j} = \left\{ \begin{array}{rcl}
 -\frac{1}{|Y_+|+\sum_{b\in Y_-} \hat{y}_b}, & \mbox{for} & j \in Y_+ \\
 -\frac{\sum_{f\in Y_+}\hat{y}_f}{(|Y_+| + \sum_{b\in Y_-}\hat{y}_b)^2}, & \mbox{for} & j \in Y_-
\end{array}\right.
\end{equation}
Since the inequality {\small $\sum_{f\in Y_+}\hat{y}_f < (|Y_+|+\sum_{b\in Y_-} \hat{y}_b)$} holds by definition, the JACLoss assigns larger gradients to foreground pixels that intrinsically balances the foreground and background classes. It is empirically works better than the cross-entropy loss or the classed balanced cross-entropy loss \cite{xie2015holistically} when segmenting small objects, such as pancreas in CT/MRI images. Similar loss functions are independently proposed and utilized in \cite{DBLP:journals/corr/MilletariNA16,DBLP:journals/corr/ZhouXSFY16}.

\section{Experimental Results and Analysis} \label{section:experiments}
{\bf Datasets:} Two annotated pancreas datasets are utilized for experiments. The first NIH-CT-82 dataset \cite{roth2015deeporgan,DBLP:journals/corr/RothLFSS16} is publicly available and contains 82 abdominal contrast-enhanced 3D CT scans. We obtain the second dataset UFL-MRI-79 from \cite{Cai2016}, with 79 abdominal T1-weighted MRI scans acquired under multiple controlled-breath protocol. For the case of comparison, 4-fold cross validation is conducted similar to \cite{roth2015deeporgan,DBLP:journals/corr/RothLFSS16,Cai2016}. Unlike \cite{DBLP:journals/corr/RothLFSS16}, no sophisticated post processing is employed. We measure the quantitative segmentation results using dice similarity coefficient (DSC): {\small $DSC=2(|Y_+\cap\hat{Y}_+|)/(|Y_+|+|\hat{Y}_+|)$}, and jaccard index (JI): {\small $JI=(|Y_+\cap\hat{Y}_+|)/(|Y_+\cup\hat{Y}_+|)$}.

{\bf Network Implementation:} Hyper-parameters are determined via model selection inside training dataset. The network that contains five Scale blocks with four CBR blocks in each Scale block produces the best empirical performance, while remaining with the compact model size ($<$ 3 million parameters). Training folds are first split into a training subset for network parameter training and a validation subset for learning hyper-parameters. Note the training accuracy as $Acc_{t}$ after model selection. We then combine training and validation subsets to further fine-tune the network until its performance on validation subset converges to $Acc_{t}$. The average time for model training is $\sim$3 hours on a single {\it GeForce GTX TITAN}.  

\begin{figure}[t!]
\begin{floatrow}
\ffigbox{%
  \includegraphics[width=5.0cm, height=3.0cm]{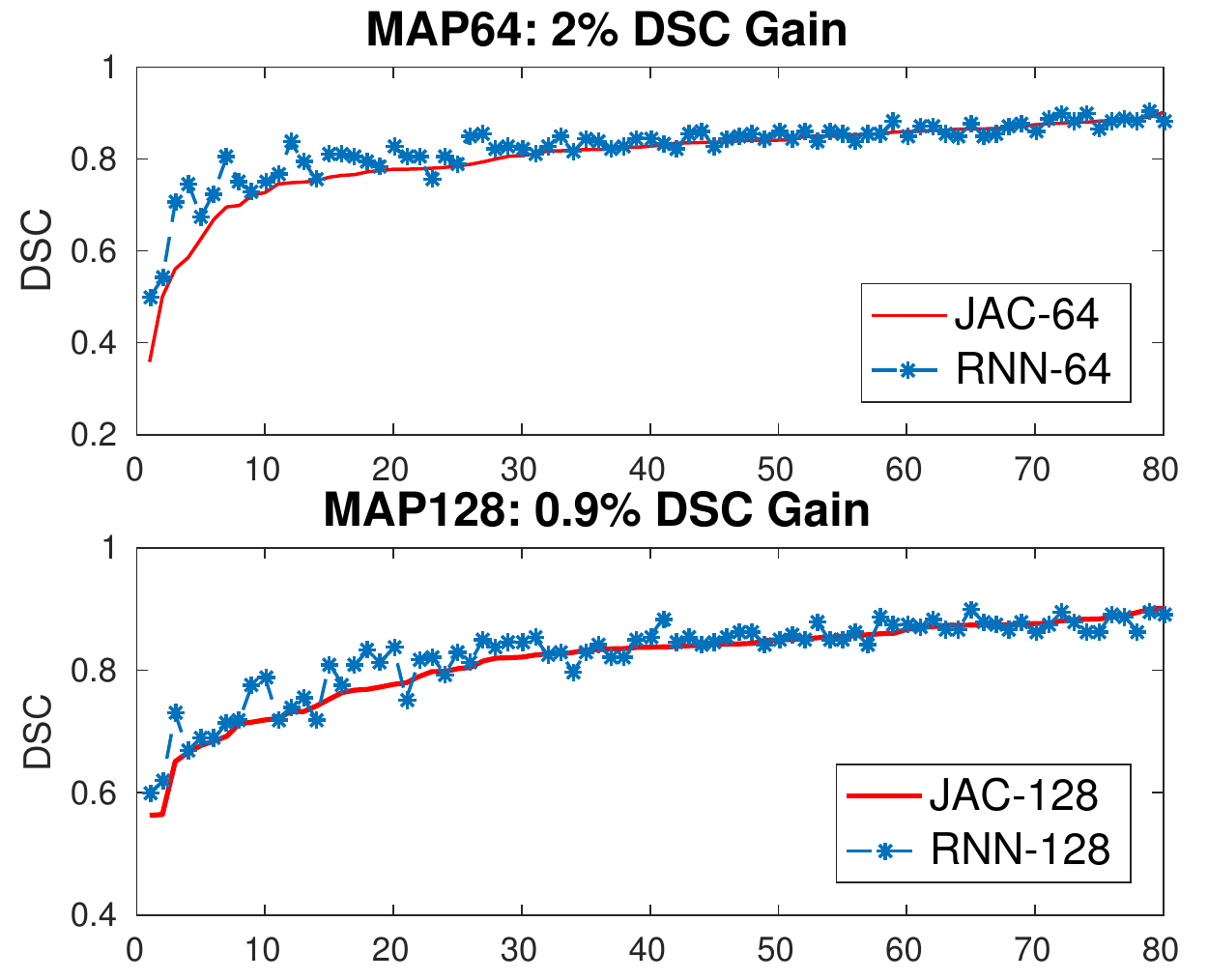}
}{%
  \caption{\small 80 cases with/without contextual learning, and sorted left to right by DSC values of JAC-models with no contextual learning. Small fluctuations among good cases are normally resulted from model updating.}%
  \label{fig:cl-improve}
}
\ffigbox{%
  \includegraphics[width=5.0cm, height=3cm]{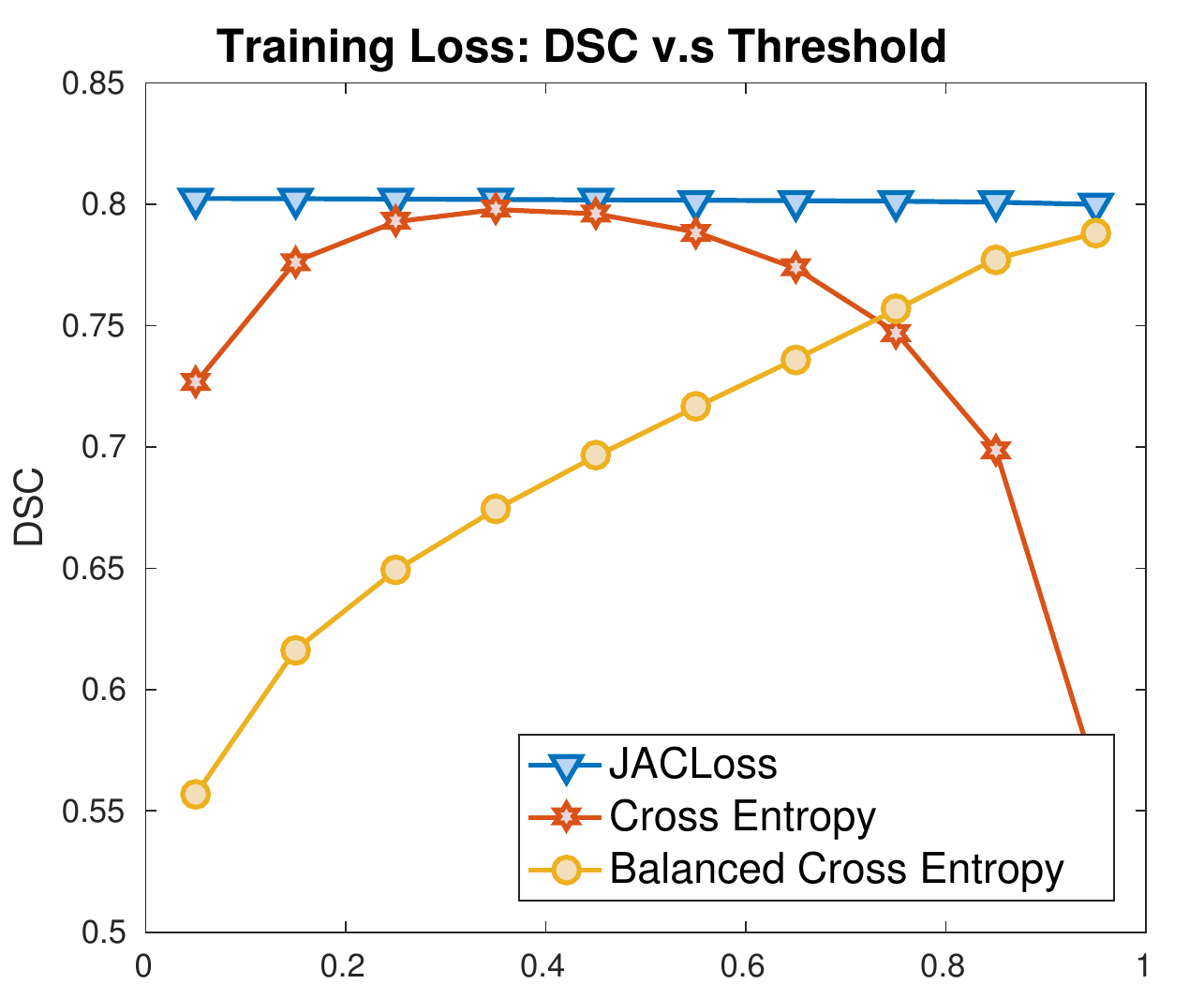}
}{%
  \caption{\small Thresholded results of models that are trained with different loss functions. The proposed Jaccard loss (JACLoss) performs stable across thresholds in range $[0.05,0.95]$.}%
  \label{fig:box-plot}
}
\end{floatrow}
\end{figure}

{\bf Analysis on Contextual Regularization:} We evaluate the proposed neural network architectures with and without contextual learning on both CT and MRI datasets. 
We first train a network of five Scale blocks with JACLoss. The number of output feature channels per convolutional layer is set as 64, and we name this network JAC-64. CLSTM contextual regularization is then applied on JAC-64's five Scale block outputs, forming a new extension of RNN-64. RNN-64 is initialized from JAC-64 and trained with enough SGD updates until convergence. 
We next investigate the performance impact on increasing the convolutional output channels from 64 to 128. Similarly, JAC-128 and RNN-128 are used to denote this variant and its contextually regularized version, respectively. From Table~\ref{tab:cpr-state-of-the-art}, RNN-enhanced deep models improve upon JAC-64/JAC-128 by 2.0\% and 0.9\% in mean DSC on NIH-CT-82. For UFL-MRI-79, RNN-64 achieves 1.8\% mean DSC gain against JAC-64. RNN-128 and JAC-128 produce the best segmentation results comparably. Fig.\ref{fig:cl-improve} further shows the segmentation performance difference statistics, with or without contextual learning. Especially, these cases with low DSC scores are greatly improved by contextual learning.

{\bf Analysis on Jaccard Loss:} Fig.~\ref{fig:box-plot} represents quantitative segmentation results of the three losses under 4-fold CV. JACLoss achieves the highest mean DSC, regardless of different segmentation thresholds. FCN or HNN outputs probabilistic image segmentation maps instead of binary masks. Thus an appropriate probability threshold is required to obtain the final binary segmentation outcomes. Na\"{i}ve cross-entropy loss assigns the same penalty on positive and negative pixels so the probability threshold should be around 0.5. Its class-balanced version gives higher penalty scores on positive pixels (due to its scarcity),  making the resulted ``optimal'' threshold at a relatively higher value. In contrast, JACLoss can push the foreground pixels to the probability of 1 while remains being strongly discriminative against background pixels. 

\begin{figure*}[t!]
	\centering
	\begin{subfigure}[b]{0.32\linewidth}
		\centerline{\includegraphics[height=1.6cm, width=3.8cm]{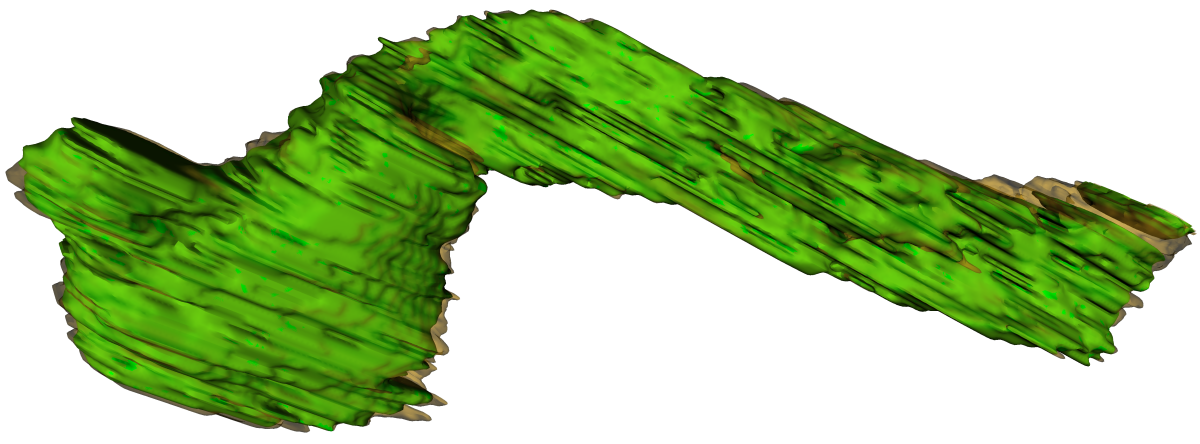}}
	\end{subfigure}
	\begin{subfigure}[b]{0.32\linewidth}
		\centerline{\includegraphics[height=1.6cm, width=3.8cm]{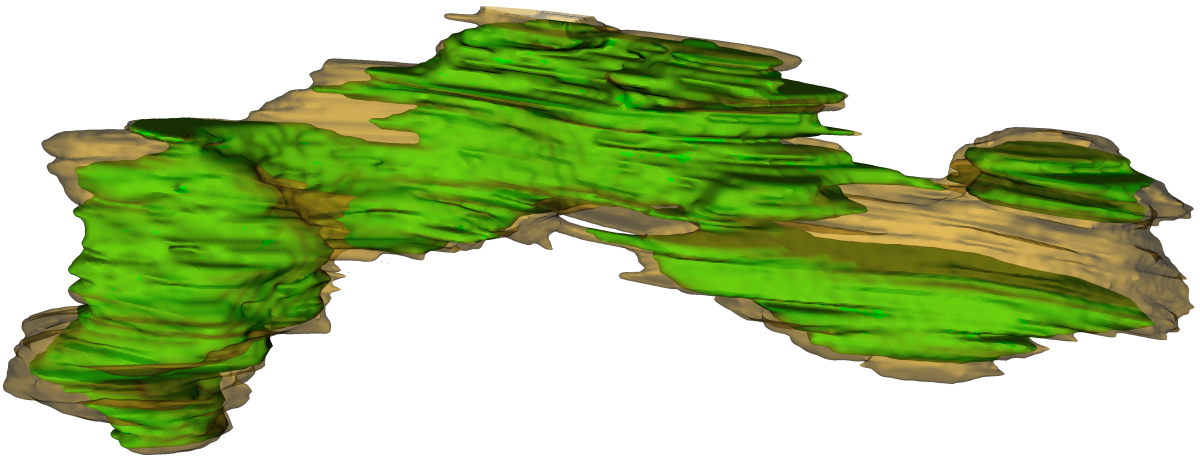}}
	\end{subfigure}	
	\begin{subfigure}[b]{0.32\linewidth}
		\centerline{\includegraphics[height=1.6cm, width=3.8cm]{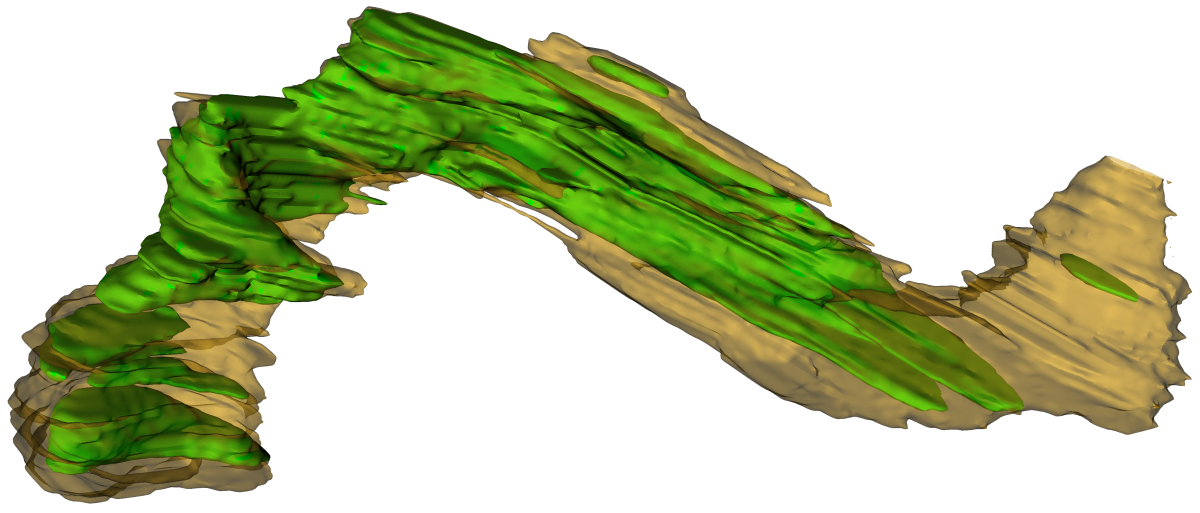}}
	\end{subfigure}		
	\caption{\small {\bf 3D visualization of pancreas segmentation results:} human annotation shown in golden and computerized segmentation displayed in green. The DSC are 90\%, 75\%, and 60\% for three examples from left to right, respectively.}
	\label{fig:3d-plot}
\end{figure*}

{\bf Comparison with the State-of-the-art Methods:} Last, we compare our pancreas segmentation models (as trained above) with the  state-of-the-art methods. Holistically-nested network \cite{xie2015holistically} (HNN) is a CNN architecture that is originally proposed for semantic edge detection. HNN has been adapted for pancreas segmentation in \cite{DBLP:journals/corr/RothLFSS16} and proved with good performance. We also implement UNet \cite{Ronneberger2015} for universal medical image segmentation problems. As a reference that {\bf HNN and UNet contains 10 times more parameters than JAC-64}, we choose to fine-tune both networks from pre-trained models. Lower layers of HNN are transferred from VGG16 while UNet parameters are transferred from the snapshot released in \cite{Ronneberger2015}. Dice similarity coefficient (DSC), and Jaccard index (JI) results computed from their segmentation outputs are reported in Table~\ref{tab:cpr-state-of-the-art}, under the same 4-fold cross validation. RNN-128 performance best on CT-82, and JAC-128 achieves the best result on MRI-79. For self-contained content, results reported in \cite{Cai2016,Roth2017,DBLP:journals/corr/ZhouXSFY16} are also included in Table~\ref{tab:cpr-state-of-the-art}. Note that our method development is orthogonal to the principles of ``coarse-to-fine'' pancreas location and detection \cite{Roth2017,DBLP:journals/corr/ZhouXSFY16}. Better performance for pancreas segmentation may be achievable with the combination of both methodologies. Fig.\ref{fig:3d-plot} displays exemplars of reconstructed segmentation results from NIH-CT-82 dataset.

\begin{table}[t!]
  \centering
  \caption{\small {\bf Comparison with the state-of-the-art methods under 4-fold cross validation:} JAC and RNN represent networks trained with JACLoss and contextual regularization, respectively. -64 and -128 represent numbers of convolutional output channels. We show dice similarity coefficient (DSC), jaccard index (JI) as mean $\pm$ standard dev. [worst, best]. The best result on CT and MRI are reported by RNN-128 and JAC-128 with bold font.}
  \label{tab:cpr-state-of-the-art}
  \scriptsize{
  \begin{tabular*}{\textwidth}{l @{\extracolsep{\fill}} cc|cc} \hline
      {\bf Method}                       & \multicolumn{2}{c}{\bf NIH-CT82} & \multicolumn{2}{c}{\bf MRI-79} \\ \hline
    			                         & DSC(\%)  & JI(\%) & DSC(\%) & JI(\%) \\ \hline 
      UNet\cite{Ronneberger2015}         & 79.7$\pm$7.6 [43.4,89.3] & 66.8$\pm$9.60 [27.7,80.7] & 79.9$\pm$7.30 [54.8,90.5] & 67.1$\pm$9.50 [37.7,82.6]  \\ 
      HNN\cite{xie2015holistically} 	 & 79.6$\pm$7.7 [41.9,88.0] & 66.7$\pm$9.40 [26.5,78.6] & 75.9$\pm$10.1 [33.0,86.8] & 62.1$\pm$11.3 [19.8,76.6]  \\
      JAC-64                             & 80.3$\pm$9.0 [35.8,90.2] & 67.9$\pm$10.9 [21.8,82.1] & 76.3$\pm$12.9 [6.30,88.8] & 63.1$\pm$14.0 [3.30,79.9]  \\
      JAC-128							 & 81.5$\pm$7.2 [56.3,90.1] & 69.3$\pm$9.50 [39.2,82.0] & \textBF{80.5$\pm$6.70} [59.1,89.4] & \textBF{67.9$\pm$8.90} [41.9,80.9]  \\
      RNN-64		                     & 82.3$\pm$6.7 [49.8,90.2] & 70.4$\pm$8.60 [33.1,82.2] & 78.1$\pm$9.40 [39.5,90.0] & 64.9$\pm$11.4 [24.6,81.8]  \\ 
      RNN-128                            & \textBF{82.4$\pm$6.7} [60.0,90.1] & \textBF{70.6$\pm$9.00} [42.9,81.9] & 80.4$\pm$6.60 [58.9,90.0] & 67.7$\pm$8.70 [41.8,81.8]  \\ \hline \hline
      Roth {\it et.al,}\cite{Roth2017}   & 81.3$\pm$6.3 [50.6,88.9] & 68.8$\pm$8.12 [33.9,80.1] & -                         & -                          \\
      Zhou {\it et.al,}\cite{DBLP:journals/corr/ZhouXSFY16} & 82.3$\pm$5.6 [62.4,90.8] & -                         & -                         & -                          \\ 
      Cai {\it et.al,}\cite{Cai2016}     & -                        & -                         & 76.1$\pm$8.7 [47.4,87.1]  & -                          \\ \hline
    \end{tabular*}
  }	
\end{table}

\section{Conclusion}
In this paper, we use a new deep neural network architecture for pancreas segmentation, via our tailor-made convolutional neural network followed by convolutional LSTM to regularize the segmentation results on individual image slices, unlike the independent process assumed in previous work \cite{Cai2016,DBLP:journals/corr/RothLFSS16,Roth2017,DBLP:journals/corr/ZhouXSFY16}. The contextual regularization permits to enforce the pancreas segmentation spatial smoothness explicitly. Combined with the proposed JACLoss function for CNN training to generate threshold-free segmentation results, our quantitative pancreas segmentation results improve the previous state-of-the-art approaches \cite{Cai2016,DBLP:journals/corr/RothLFSS16} on both CT and MRI datasets.

\bibliographystyle{splncs}
\bibliography{ref}

\end{document}